\pdfoutput=1

\documentclass[11pt]{article}

\usepackage[]{acl} 
\usepackage{times}
\usepackage{latexsym}
\usepackage[T1]{fontenc}
\usepackage[utf8]{inputenc}
\usepackage{microtype}
\usepackage{inconsolata}
\usepackage{graphicx}
\usepackage{float}
\usepackage{booktabs}

\title{Developing a Mixed-Methods Pipeline for Community-Oriented Digitization of Kwak'wala Legacy Texts}

\author{
 \textbf{Milind Agarwal\textsuperscript{1}},
 \textbf{Daisy Rosenblum\textsuperscript{2}},
 \textbf{Antonios Anastasopoulos\textsuperscript{1}},
\\
 \textsuperscript{1}George Mason University,
 \textsuperscript{2}University of British Columbia,
\\
 \small{
   \textbf{Correspondence:} \href{mailto:magarwa@gmu.edu}{magarwa@gmu.edu}
 }
}

\begin{document}
\maketitle
\begin{abstract}
Kwak'wala is an Indigenous language spoken in British Columbia, with a rich legacy of published documentation spanning more than a century, and an active community of speakers, teachers, and learners engaged in language revitalization.  Over 11 volumes of the earliest texts created during the collaboration between Franz Boas and George Hunt have been scanned but remain unreadable by machines. Complete digitization through optical character recognition has the potential to facilitate transliteration into modern orthographies and the creation of other language technologies. In this paper, we apply the latest OCR techniques to a series of Kwak'wala texts only accessible as images, and discuss the challenges and unique adaptations necessary to make such technologies work for these real-world texts. Building on previous methods, we propose using a mix of off-the-shelf OCR methods, language identification, and masking to effectively isolate Kwak'wala text, along with post-correction models, to produce a final high-quality transcription.\footnote{Relevant code and data resources are available \href{https://github.com/magarw/kwakwala}{here}.}

\end{abstract}

\section{Introduction}

\begin{figure*}
    \centering
    \includegraphics[width=0.8\linewidth]{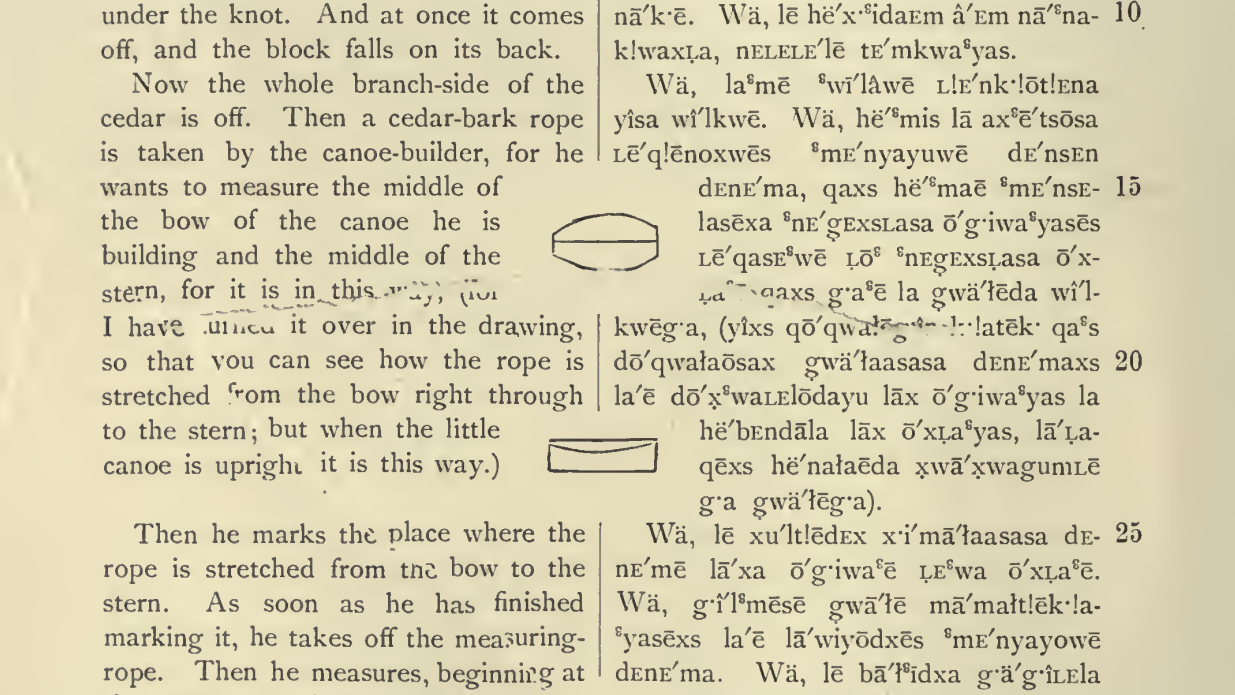}
    \caption{Example two-column text from the Kwakiutl of Vancouver Island (1909) collection. Notice the abundance of inline figures in this text that interfere with Google Vision's OCR pipeline.}
    \label{fig:two-col-figure}
\end{figure*}

In this work, we focus on the Kwak’wala language (Wakashan, ISO 639.3 kwk), spoken on Northern Vancouver Island, nearby small islands, and the opposing mainland. Kwak’wala and several other Indigenous languages in this region have over a century of of legacy documentation created by early anthropologists, primarily in orthographies developed by Franz Boas to capture complex and typologically unusual phonetic and phonological inventories \cite{himmelmann1998documentary,grenoble2005saving}. Kwak’wala, for example, has 42 consonant phonemes represented with a selection of characters from the North American Phonetic Alphabet (cousin to the IPA), and over 13 possible vowel pronunciations represented with a heavy dose of diacritics and digraphs in all its scripts.  During the first half of the 20th-century, scripts such as these were created and used by ethnologists, researchers, and collectors to transcribe the languages spoken in communities across North America. Between 1897 and 1965, an extensive series of texts in Indigenous languages was published by the United States Bureau of American Ethnology (BAE, now the Smithsonian). The collaboration between Franz Boas and George Hunt generated 11 volumes of published texts over 50 years, as well as extensive unpublished documentation. This script is difficult for anyone  to read, amplifies phonetic complexity, and is primarily considered a legacy script, limiting access only to a few. However, many precious documents with detailed information of cultural value, were created in this script (see Figure \ref{fig:two-col-figure}), necessitating their accurate digitization and transliteration into modern Kwak'wala writing systems. Note that while Kwak’wala is classified as an endangered language with most first-language speakers over the age of 70, it has  thriving language revitalization programs focused on creating new speakers among children and adults. Research progress for Kwak’wala and its three scripts (U’mista in the Northern communities, SD-72 in the Southern communities, and the legacy Boas-Hunt script) is urgent to better support revitalization and educational efforts led by community members. Currently, Kwak’wala, like many other `low-resource' endangered and Indigenous languages, lags behind in the number of available computational tools \cite{agarwal-anastasopoulos-2024-concise}. 

To remove this disadvantage and enable greater online community participation, in our project, we focus on digitization of valuable Kwak'wala texts, prioritized according to community needs, to enable building tools such as word processing, speech to text, predictive typing, etc. We create these resources by applying existing optical character recognition and language identification techniques, and making necessary modifications to suit them to Kwak'wala.  We use grapheme-to-phoneme technology \cite{pine-etal-2022-gi22pi} to transliterate texts into the U'mista orthography, one of two community-preferred modern Kwak’wala writing systems. A draft of the 1921 Boas-Hunt text produced through a previous collaboration was distributed to 50 community and academic experts for review, and the feedback we gathered through surveys and conversations informed our production of a second draft PDF for publication and distribution. This feedback assisted us in prioritizing highly-valued elements of the texts which had originally been overlooked or erased through the process, such as the text-referenced line numbers cited by Boas in his dictionary and grammar, creating an analog concordance and networking these Kwak'wala texts into the prototypical `Boasian trilogy'. This research, conducted in consultation with community-based language programs and guided by community priorities, will greatly increase access to culturally significant documents, thus empowering the community to draw on these resources to propagate the language and culture to future generations \cite{Lawson_2004}.

\section{Data}

We focus our effort on digitizing five  books that include Kwak'wala text and, often, parallel translations in English. We chose these books due to their similar fontfaces, clean layout, typed content (as opposed to handwritten), and high-quality scans. 

\begin{enumerate}
    \item \textbf{Jesup Volume 5, Part 1 (Franz Boas and George Hunt, January 1902)}: This 280 page book is part of the Jesup North Pacific Expedition publication series and contains an anthology of Kwak'wala texts in Hunt-Boas orthography. The book primarily contains  dictated Kwak'wala  texts (with parallel running English translation), and an appendix with  grammatical information, stems, vocabulary, and traditional songs sung by Kwak'wala communities \cite{boas1902-p1}.
    \item \textbf{Jesup Volume 5, Part 2 (Franz Boas, December 1902)}: This 144 page book is mostly formatted similarly to Volume I, but this particular volume doesn't contain interlinear text, and instead has an abundance of monolingual single-column Kwak'wala prose \cite{boas1902-p2}. 
    \item \textbf{Jesup Volume 5, Part 3 (Franz Boas, 1902)}: This is the third and final part of Volume 5, and is formatted similarly to Part II. It also contains a substantial appendix with vocabulary and stems \cite{boas1902-p3}. 
    \item \textbf{Jesup Volume 10 (Franz Boas and George Hunt, 1906)}: This book contains valuable texts from the North Pacific Expedition in Kwak'wala and Haida (Masset dialect) languages. For the purposes of this project, we use only the first part of the first 282 pages of this book that contains the Kwak'wala texts \cite{boas1906}. 
    \item \textbf{The Kwakiutl Of Vancouver Island Volume II (Franz Boas, 1909)}: This book contains valuable texts in Kwak'wala on wood-working, weaving, hunting, fishing, clothing, measurements etc. Most of the descriptions are in English, with plenty of inline figures (that disrupt the layout extraction of the OCR), but there are also tens of pages of Kwak'wala dictated text (with parallel running English). The book alternates between a single and double column layout \cite{boas1909}. 
\end{enumerate}

\section{Related Work}
Optical character recognition (OCR) is a multi-label classification problem, where a patch of pixels is shown to an OCR model, and its task it to classify it into one of \textit{n} classes (usually the alphabet + punctuation). When extended to entire pages or documents containing textual material, this can allow us to digitize previously inaccessible materials. Since it is crucial for digitization of manuscripts, linguistic field notes etc., it is widely used in the humanities to render such texts accessible to researchers and to language community members \cite{6-DBLP:journals/corr/ReulSP17,rijhwani-etal-2021-lexically,rijhwani-etal-2020-ocr}. 
 
 This technique has, over time, developed into a discipline, with many excellent surveys written covering the technical and applied aspects of OCR \cite{agarwal-anastasopoulos-2024-concise,10.1145/3453476,10.1145/3476887.3476888,9151144,hedderich-etal-2021-survey}. Today, many open-source (Tesseract and Ocular) and commercial systems (Google Vision and Microsoft OCR) exist for OCR and they can extract text from most images quite effectively, as long as they are in a language it has seen during training  \cite{17-DBLP:conf/icdar/Smith07,18-blecher2023nougat,19-DBLP:conf/acl/Berg-KirkpatrickDK13}. Several research efforts before have tried to address the lack of resources in certain indigenous languages using OCR to create machine-readable texts such as Central Quechua \cite{cordova-nouvel-2021-toward} and Akuzipik \cite{hunt-etal-2023-community}.

\section{Methodology}

\subsection{First-Pass OCR}
 Google Vision is a well-maintained modern OCR tool that tends to work well on Latin/Roman orthographies and their extensions \cite{DBLP:conf/icdar/FujiiDBHP17,rijhwani-etal-2020-ocr}. Additionally, since our collections are composed of multilingual texts, it is vital to use a tool that can handle multilinguality within documents. It is a paid (per page) service at the rate of \$1.25/1000 pages, but the first 1000 pages every month are free. Since our project and its digitization was conducted over several months, we did not incur any first-pass OCR charges. Open-source alternatives like Ocular or Tesseract may also be used for OCR, especially when data restrictions require local processing, instead of sending data through APIs to Google servers. However, note that they require manual training, computational expertise, preparation of training and evaluation data, and have a higher learning curve \cite{17-DBLP:conf/icdar/Smith07,berg-kirkpatrick-etal-2013-unsupervised}  .

\subsection{Language Identification}
We use language identification (langID) to distinguish English from Kwak'wala as proxy for structure identification in our collections. LangID is also extremely important to enable masking of non-Kwak'wala text. To the best of our knowledge, no off-the-shelf language identification model supports Kwak'wala in the Hunt-Boas orthography. So, we use fastText as it allows easily training on custom data from scratch on CPU \cite{joulin-etal-2017-bag,agarwal-etal-2023-limit}. Our final model, trained on first-pass Kwak'wala and English texts (binary model, default fastText parameters, 1000 sentences per language), achieves a sentence-level accuracy of 99.84\%. This model is applied on each page's bounding boxes, which allows us to reorganize text with improved layout.

\subsection{Masking}
The texts are diversely formatted, and contain additional information in illustrations, figures, line numbers and the like. Since the post-correction model (see Section \ref{subsec:post-correction}) is trained to correct Kwak'wala text alone, we quickly realized that real-world digitization projects like ours require the development of a masking pipeline. Additionally, masking is preferable as post-OCR correction models are best trained for a single language, and English first-pass OCR quality is often extremely good without requiring post-OCR correction. Following the first-pass OCR, we apply a masking layer that temporarily hides/masks all English text (as labeled by the langID model), numbers, and certain punctuation like parentheses, that were impacting subsequent steps in the pipeline. This allows us to isolate, to the best ability of the language identification model and based off our overall structural cropping, the first-pass text in Kwak'wala that needs post-correction. For each line, token-level indices of the masked tokens are stored in a separate file at this stage. This allows us to track exactly what tokens were masked so we can reintroduce them in the same spots after post-correction.

\subsection{Post-Correction}
\label{subsec:post-correction}
Post-correction can allow us to automatically correct errors in very low-resource OCR settings, by training a correction model on a small sample of first-pass and reference text pairs \cite{kolak-resnik-2005-ocr, dong-smith-2018-multi}. The post-correction model has a multi-source neural architecture, based on \citet{rijhwani-etal-2021-lexically},  which has been shown to reduce character error rates by 32–58\%. We use the model from this paper directly for post-correction, with the weighted finite-state transducer setting for lexical induction turned off, as it was shown in \citet{rijhwani-etal-2021-lexically} not to improve Kwak'wala post-correction in contrast to other low-resource languages. This is likely due to the polysynthetic structure of Kwak'wala words, leading to low lexical frequency of any one token. We train the model from scratch on the labeled Boas-Hunt dataset shared in the paper, with pretraining conducted on the unlabeled first-pass OCR outputs for the collection. We first replicated the character error rate results from the original paper to ensure reliability of the model. Then we applied our trained model to our test text. The unmasked Kwak'wala text from the previous stage is fed line-by-line to the post-correction model to obtain post-corrected Kwak'wala text.

\subsection{Reconstruction}
Next, we reinsert the masked tokens (English text, punctuation, line numbers in the margins, etc.) into the post-corrected sentence  at the appropriate indices. This gives us the final reconstructed multilingual output, along with crucial indexical cross-referencing information such as page and line numbers. At this stage, the Kwak'wala text is also transliterated into the desired modern orthography (ex. U'mista or SD-72) using grapheme-to-phoneme conversion to allow for better readability and accessibility of the text. 

\subsection{Evaluation}
We compare the reconstructed output to gold reference texts to evaluate the digitized texts' quality. We do this for two books at two levels:
\begin{itemize}
    \item \textbf{Textual Errors}: To capture textual errors, such as misspellings, missing diacritics, tokenization etc., we use Character Error Rate (CER). This is a popular metric to understand character-level variations and error distributions in the output text, as compared to the gold-reference.  For morphologically complex and polysynthetic languages like Kwak'wala, CER is a much better metric than word-level scores because a large amount of vocabulary would be unseen at test-time \cite{rijhwani-etal-2023-user}.

    \item \textbf{Structural Errors}: We use the metric from \citet{structural-zoning} that measures insertion, deletion, and maximal move operations required across the output page to  make it identical with the reference text. A weighted sum of these operations gives us the overall error, allowing us to quantify the structural quality of out outputs, and we normalize it to be between 0-100, with less being better.
\end{itemize}
Gold reference pages are created by inspecting the post-corrected output, comparing it with the source image, and manually correcting any errors. This is the most expensive part of the overall process and requires expertise in the language. So, for the moment, we evaluate on a few representative sample pages for two books.  We showcase these results in Table \ref{tab:main_results}, where we can observe a 50\% decrease in character error rate and 87.5\% reduction in structural error with our pipeline of language identification, masking, and automatic post-correction.

\begin{table}
    \centering
    \begin{tabular}{ccccc}
    \toprule
        &  \multicolumn{2}{c}{Jesup 5.1, 1902} & \multicolumn{2}{c}{Kwakiutl, 1909}  \\
        &  CER & SER & CER & SER \\
    \midrule
    First Pass & 0.43 & 25 & 0.33 & 18\\
    Corrected & 0.18 & 2 &  0.15 & 3 \\
    \bottomrule
    \end{tabular}
    \caption{For both books, we find that using our pipeline greatly reduces not only textual errors (CER) but also greatly improves the layout and structure (SER)}
    \label{tab:main_results}
\end{table}

\section{Conclusion}
We apply the latest OCR techniques to a series of previously undigitized Kwak'wala texts, and demonstrate the challenges and unique adaptations necessary to make OCR work for real-world texts and collections. We propose using a mix of off-the-shelf OCR methods, language identification and masking to effectively isolate Kwak'wala text, and post-correction models to produce a  high-quality transcription.  We plan to disseminate the digitized documents directly to the community members. Additionally, with consent of the community partners and data annotators, we plan to share the digitized and transliterated text (in three orthographies) with the data hosting institutions, such as the American Philosophy Society and Columbia University Rare Books and Special Collections, where a large collection of Boas-Hunt manuscripts have recently been digitized \cite{rbml2023georgehunt}.  We hope to explore ways that this work in improving OCR for Kwak’wala and developing reliable digitization workflows for legacy texts can be transferable to other legacy orthographies, directly benefitting other language communities.

\section*{Limitations}
Since our contribution type is best suited to a short paper, at the moment, we did not include more extensive benchmarking for language identification. As we continue to work with our language community collaborators, we will continue to add more gold reference texts for comparison and better evaluation of the transcriptions.  

\section*{Ethics Statement}
Though they derive from material in the public domain, the first-pass, gold reference texts, and corrected transcriptions of the selected Kwak'wala texts will only be released publicly with the consent of the language community members. An ethical implication of this work is that it will allow for  more sustainable and equitable work in language resource creation and natural language processing, under the guidance of the language community members and their immediate and long-term needs for effective Kwak'wala revitalization.

\section*{Acknowledgments} This work was generously supported by the National Endowment for the Humanities under award PR-276810-21, George Mason University's Doctoral Research Scholars Award 2024-25 and the Stanford Initiative on Language Inclusion and Conservation in Old and New Media (SILICON) Practitioners 2024-25 Award.  The authors are also grateful to the anonymous reviewers for their valuable suggestions, feedback, and comments. 

\bibliography{custom,computel}

\end{document}